\documentclass[sigconf]{acmart}
\usepackage{balance}

\AtBeginDocument{%
	\providecommand\BibTeX{{%
			\normalfont B\kern-0.5em{\scshape i\kern-0.25em b}\kern-0.8em\TeX}}}

\renewcommand\footnotetextcopyrightpermission[1]{}
\setcopyright{none}
\acmDOI{}

\begin{document}
	\fancyhead{}
	
	\title{Dynamic Future Net: Diversified Human Motion Generation}
	
	% \affiliation{%
	\author{Wenheng Chen}
	\affiliation{%
		\institution{NetEase Fuxi AI Lab}
		%   \city{HangZhou}
		%   \country{China}
	}
	\email{chenwenheng@corp.netease.com}

	\author{He Wang}
	\affiliation{%
		\institution{University of Leeds}
		%   \country{China}
	}
	\email{H.E.Wang@leeds.ac.uk}

	\author{Yi Yuan}
	\authornote{Corresponding author}
	\affiliation{%
		\institution{NetEase Fuxi AI Lab}
		%   \postcode{78229}
	}
	\email{yuanyi@corp.netease.com}
	
	\author{Tianjia Shao}
	\affiliation{%
		\institution{State Key Lab of CAD\&CG, Zhejiang University}
		%   \country{China}
	}
	\email{tjshao@zju.edu.cn}

	\author{Kun Zhou}
	\affiliation{\institution{State Key Lab of CAD\&CG, Zhejiang University}}
	\email{kunzhou@acm.org}

	\renewcommand{\shortauthors}{Chen and Wang, et al.}
	
	\begin{abstract}
		Human motion modelling is crucial in many areas such as computer graphics, vision and virtual reality. Acquiring high-quality skeletal motions is difficult due to the need for specialized equipment and laborious manual post-posting, which necessitates maximizing the use of existing data to synthesize new data. However, it is a challenge due to the intrinsic motion stochasticity of human motion dynamics, manifested in the short and long terms. In the short term, there is strong randomness within a couple frames, e.g. one frame followed by multiple possible frames leading to different motion styles; while in the long term, there are non-deterministic action transitions. In this paper, we present \textit{Dynamic Future Net}, a new deep learning model where we explicitly focuses on the aforementioned motion stochasticity by constructing a generative model with non-trivial modelling capacity in temporal stochasticity. Given limited amounts of data, our model can generate a large number of high-quality motions with arbitrary duration, and visually-convincing variations in both space and time. We evaluate our model on a wide range of motions and compare it with the state-of-the-art methods. Both qualitative and quantitative results show the superiority of our method, for its robustness, versatility and high-quality.
	\end{abstract}
	
	
	
	% \keywords{human motion, neural networks, generative models}
	
	\begin{teaserfigure}
		\centering
		\includegraphics[width=0.9\textwidth]{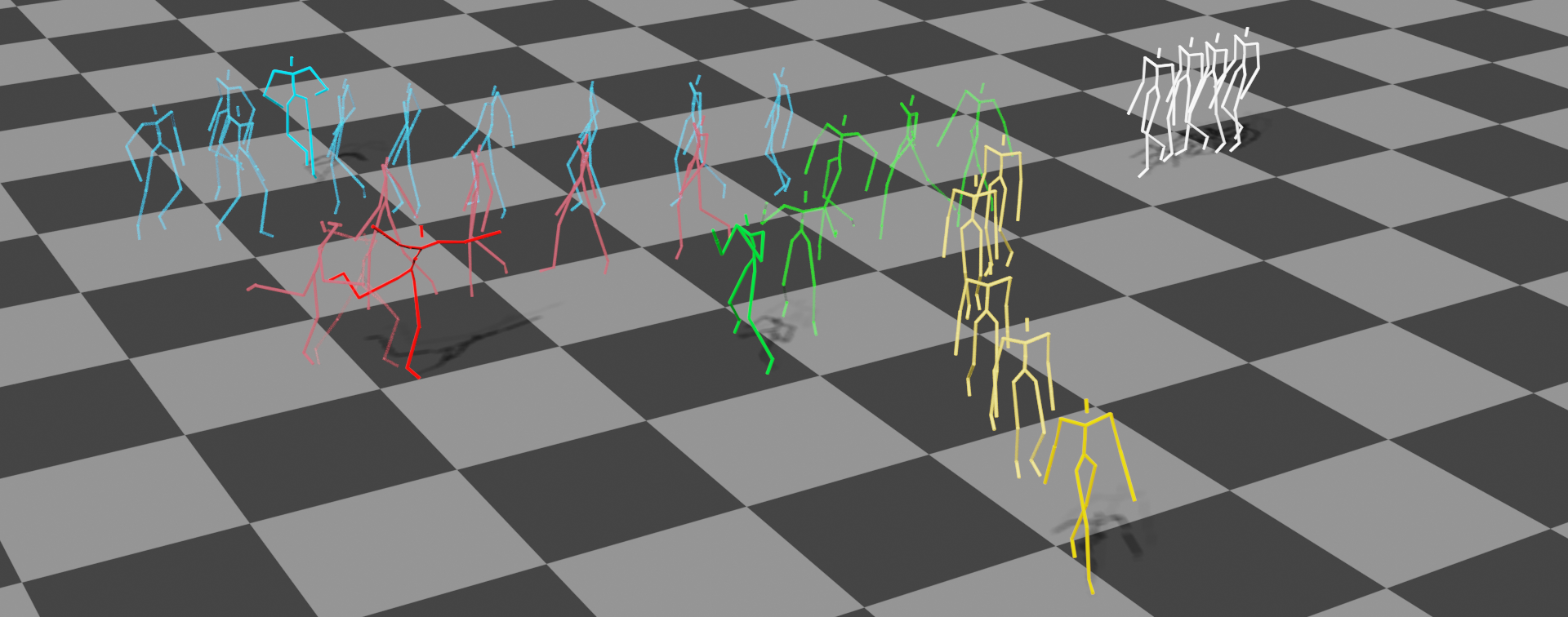}
		\caption{Given a 20-frame walking motion pre-fix (white), our model can generate diversified motion: walking (yellow), walking-to-running (blue), walking-to-boxing (green), and walking-to-dancing (red), with arbitrary duration. The corresponding animation can be found in teaser.mp4 in supplementary video.}
		\label{fig:teaser}
	\end{teaserfigure}
	
	\maketitle
	
	\fancyfoot{}
	\thispagestyle{empty}

	\section{Introduction}
	Modeling natural human motions is a central topic in several fields such as computer animation, bio-mechanics, virtual reality, etc, where high-quality motion data is a necessity. Despite the improved accuracy and lowered costs of motion capture systems, it is still highly desirable to make full use of existing data to generate diversified new data. One key challenge in motion generation is the dynamics modelling, where it has been shown that a latent space can be found due to the high coordination of body motions \cite{holden2016deep,wang2019spatio,sythnphy}. However, as much as the spatial aspect is studied, dynamics modelling, especially with the aim of diversified motion generation, still remains to be an open problem.
	
	Human motion dynamics manifest several levels of short-term and long-term stochasticity. Given a homogeneous discritization of motions in time, the short-term stochasticity refers to the randomness in next one or few frames (pose-transition); while the long-term one refers to the random in the next or few actions (action-transition). Tradition methods model them by Finite State Machines with carefully organized data \cite{MGplusplus}, which have limited model capacities for large amounts of data and require extensive pre-processing work. New deep learning methods either ignore them \cite{holden2016deep} or do not explicitly model them \cite{wang2019spatio}. Very recently, dynamics modelling for diversified generation has just been investigated \cite{wang2019combining}, but only from the perspective of the overall dynamics, rather than the detailed short/long term stochasticity.
	
	In this paper, we propose a new deep learning model, Dynamic Future Net, or DFN, for automatic and diversified high-quality motion generation based on limited amounts of data. Given a motion, we assume that it can be discretized homogeneously in time and represented by a series of posture and instantaneous velocities. Following the observation that it is easier to learn the dynamics in a natural motion latent space \cite{holden2016deep}, we first embed features in the data space into a latent space. Next, DFN learns explicitly the \textit{history}, \textit{current} and \textit{future} state given any time, where we also model several conditional distributions for the influences of history and future state on the current state. The state-wise bidirectional modelling (extending into both the past and future) separates DFN from existing methods and endows us with the ability of modelling the short-term (next-frame) randomness and long-term (next-action) randomness. Last, for inference purposes, we propose new loss functions based on distributional similarities as opposed to point-wise estimation \cite{wang2019spatio,zhou2018autoconditioned}, which captures the dynamics accurately but also keep the randomness that is crucial for diversified motion generation.
	
	We show extensive experimental results to show DFN's robustness, versatility and high-quality. Unlike existing methods which have to be trained on one type of motions a time \cite{wang2019combining,zhou2018autoconditioned}, DFN can be trained using both single type of motions or mixed motions, which shows DFN's ability to capture multi-modal dynamics and therefore its versatility in diversified motion generation. Visual evaluation shows that DFN can generate high-quality motions with different dynamics.
	
	In summary, our formal contributions include:
	
	\begin{enumerate}
		\item a new deep learning model, Dynamic Future Net, for automatic, diversified and high-quality human motion generation.
		\item a new dynamic model that captures the transition stochasticity of the past, current and future states in motions.
		\item insights of the importance of both short-term and long-term dynamics in human motion modelling.
	\end{enumerate}

	\section{Background and related works}
	\subsection{Human pose and motion embedding} Given a human motion sequence, it is useful to find the low dimension representation of the whole sequence. Holden et al. \cite{holden2015learning} for the first time use a convolution neural network to project the entire sequence to a low dimensional embedding space. Using this more abstract representation, one can blend motions or remove noises from corrupted motions. In \cite{holden2016deep,holden2017phase}, the authors further make use of the power of the learned motion manifold and decoder to synthesize motions with constraints. Another important application of motion embedding is motion style generation \cite{du2019stylistic}, in which the embedding code can be tuned to matched the desired style. Although modeling a motion sequence with auto-encoders is straightforward, how it can model the dynamics of human motion is not clear. In \cite{kundu2019unsupervised}, the authors model a motion sequence as a trajectory in pose embedding manifold, then use a bidirectional-RNN to encode the trajectory to model the dynamics of the trajectory, which can improve motion classification results. Moreover, they designed a graph-like network to better represent the human body components.
	
	The existing methods focus on the embedding of the poses and dynamics. However, they do not explicitly model the distributions of these latent variables which governs the stochasticity of the dynamics. In this paper, we go a lever deeper and learn the latent variable distributions for the embedded poses and dynamics.

	\subsection{Deterministic human motion prediction and synthesis}
	In the effort of modelling motion dynamics, many methods employ deterministic transitions \cite{hu2019predicting,guo2019human,butepage2017deep,bartoli2018context,fragkiadaki2015recurrent,harvey2018recurrent,Co-attention,hernandez2019human,mao2019learning,li2018convolutional}, especially in human motion prediction or generation. They either focus on short term dynamics modeling or spatial-temporal information of the overall dynamics. In \cite{zhou2018autoconditioned}, the authors propose a training technique for RNN to generate very long human motions. Although this technique solves the problem of the freezing phenomena of RNN, their model is deterministic, which makes the training difficult: given a past state, if multiple possible future motions are present in the data, the network will average them, which is a common problem in many human motion prediction methods.
	
	%%AI4Animation % language to pose % music to pose
	One solution to this problem is to introduce control signals \cite{starke2019neural,habibie2017recurrent}. They design several networks and make the character to follow a given trajectory in real-time. In \cite{peng2018deepmimic}, the control signal becomes the 3d human pose predicted by neural nets as a reference for an agent to imitate. In \cite{ahuja2019language2pose}, the authors co-embed the language and corresponding motions to a share manifold, ignoring the fact that language-to-motion is a one-to-many mapping. Even with a specific control signal, like 2D human skeleton, one can still expect that there are different motions or different pose corresponding to the same control signal \cite{li2019generating}, essentially indicating the multi-modality nature of human motion dynamics.
	
	Different from the existing methods, our paper focuses on the explicit modelling of the multi-modality nature of motion transitions in human motions. Further, we also aim to learn the stochastiity in those transitions.
	
	\subsection{Stochastic human motion synthesis}
	%Nutan chen

	In \cite{wang2019combining} the authors combine RNNs and Generative Adversarial Networks (GANs) to generate stochastic human motions. They use a mixture density layer to model the stochastic property, and use an adversarial discriminator to judge whether the generated motion is natural or not. In MoGLow \cite{henter2019moglow}, the authors for the first time use normalizing flow to directly model the next frame distribution. One advantage of this method is that it can capture complex distributions without learning an apparent latent space. Given the same initial poses and the same control signals or constraints, the model still generate different motion sequences. Chen et al. \cite{chen2016dynamic} combine Dynamic motion primitive and variational Bayesian filter to model the human motion dynamics. They show that the latent representations are self-clustering after training. However, in the transitions, it needs the information of the whole sequence, which separates it from being a pure generative model.
	
	% One possible way to ensure the motion to be natural is combining physic or directly model the human muscle \cite{wei2011physically,peng2018deepmimic,DReCon,muscle,SFV}. But this line of method relies on real data or physic engine, which makes it hard to generalize to non-realistic data like cartoon animation, so we do not follow this method in our paper
	
	Our method differs from existing approaches in its treatment in the relations between the past, current and future states of human motions. Unlike the aforementioned methods, we explicitly model the current state based on both the past and the future. Also, we further model their randomness in the latent space that captures the transition multi-modality.
	
	\subsection{Stochastic RNN model}
	%% General sequence generative model VRNN, SRNN, Deep kalman filter, bayseian filter
	%% Especially:
	%% Planet, TDVAE, Z-forcing
	Modelling the stochasticity in time-series data has been a long-standing problem, such as music, hand writing and human voice \cite{vanwavenet,rnnseq,huang2018music}. The VRNN \cite{chung2015recurrent} for the first time combines Variational AutoEncoder (VAE) \cite{VAE} and recurrent neural networks for this purpose. Later in \cite{hsu2017unsupervised}, the authors disentangle the latent variables of the observation and the dynamics, with the observation latent being used to recover the full observation information, and the dynamic latent capturing the dynamics. 
	
	A key modelling choice in stochastic RNN models is the relations between the past, current and future. In early work, the posterior of current state is inferred from the past information, which makes it lack the ability to foresee the future. In \cite{goyal2017z,bayer2014learning,serdyuk2018twin}, the authors show that the performance can be improved by incorporating the future state with a backward RNN in the inference stage.  In \cite{gregor2018temporal}, the authors design a model that can go beyond step-by-step modelling, and predict multiple steps up to a given horizon. Similar effort is also made in reinforcement learning, where the reward function takes the discounted future reward into consideration \cite{gregor2018temporal,Hafner2020Dream}. In \cite{hafner2019learning}, the authors went further and designed a model that can predict multiple future scenarios, then choose the one with highest predicted reward from all the possibilities. 
	
	We observed that human motions follow a similar philosophy: the current state is a result of the past motion but also a particular choice for a certain planned future. Our research is inspired by Stochastic RNN models but focuses on human motion transition stochasticity. 
	
	\section{Method Overview}
	Our method takes a homogeneous series of human pose representations as input. This representation contains the 3D joint coordinates relative to the root, and the root translation velocity over ground plane and the rotation velocity around the y-axis. We propose the Dynamic Future Net to model the motion dynamics as a future-guided transition and generate random natural human motions that transit between different actions.
	
\begin{figure}[h]
		\centering
		\includegraphics[width=0.7\linewidth]{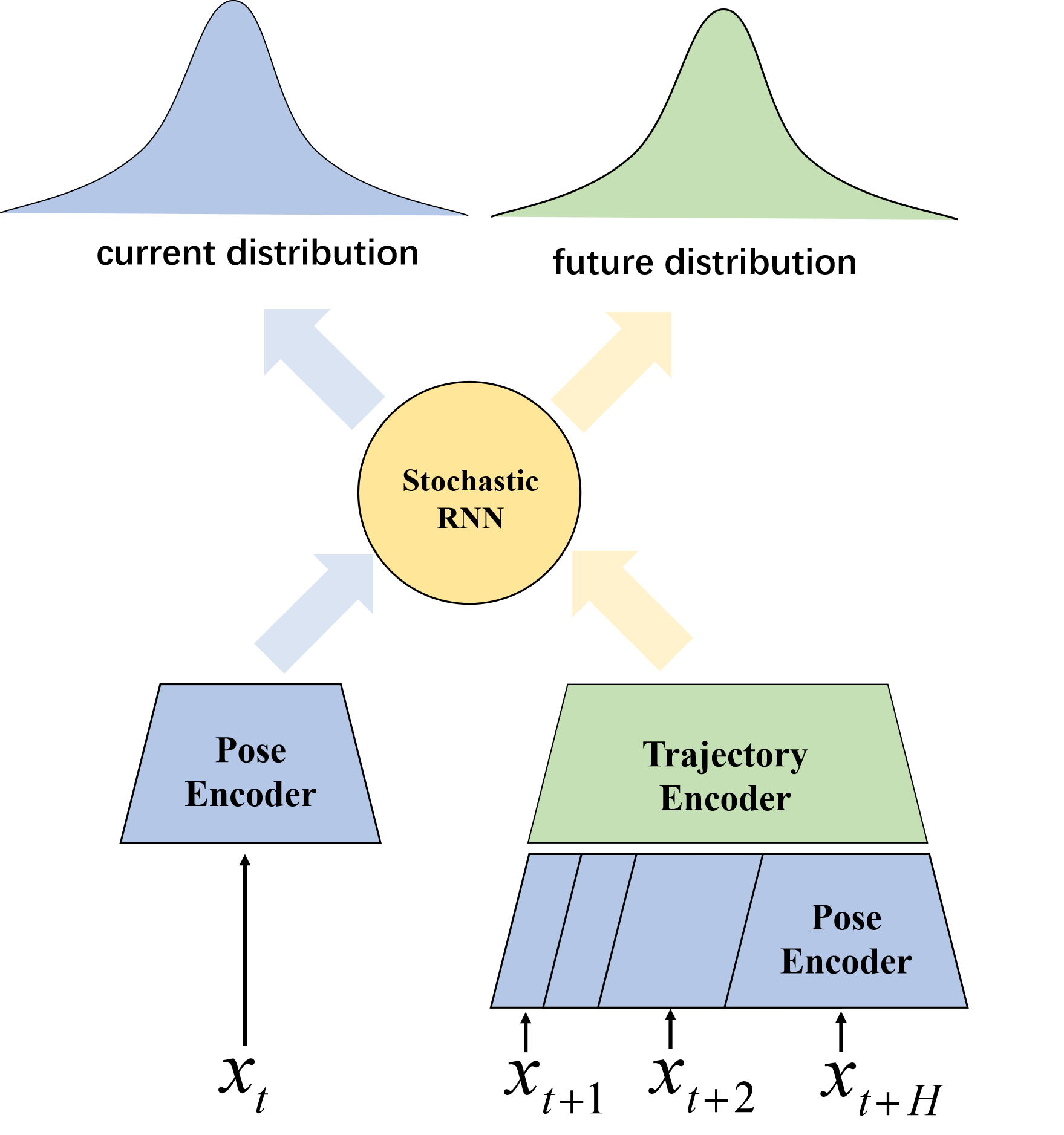}
		\caption{Overview of the proposed Dynamic Future Network. In the learning process, the network take human motion sequence as input and predict the long term distribution and the short term(next frame) distribution.}
		
		\label{fig:pipeline}
	\end{figure}

	As illustrated in Fig.~\ref{fig:pipeline}, DFN is composed of three modules, a pose encoder, a pose trajectory encoder and a stochastic latent RNN. The pose encoder (Section 5.1) maps the high dimensional human pose to a latent space while the pose trajectory encoder (Section 5.2) embeds the trajectory in the latent pose space into a code. Such compact representations of pose sequences can facilitate the learning process~\cite{kundu2019unsupervised}. As a key module, the stochastic latent RNN (Section 5.3) deploys a stochastic latent state and a deterministic latent state to learn two latent distributions for the pose-embedding and the future trajectory embedding. Such explicit learning of two different latent distributions on the one hand forces the model to learn strong temporal correlation and on the other hand generates motions with varied and natural transition. During inference we combine the past, current and future state to infer the current latent state distribution, and we combine the past and future to infer the future latent state distribution. In the generation stage, unlike existing methods \cite{goyal2017z} where the current state is generated from the past state only, we first generate the future state and combine it with the past state to generate the current latent state prior, from which we sample the current latent state then decode it to the pose-embedding and recover the current pose and velocity. We regard this process as a self-driving motion generation process guided by the envisioned dynamic future. In this way, the model can learn and generate rich and varied natural motions.
	
	\section{Data preparation}
	We train our model on the CMU Human motion capture dataset. As the skeletons in the origin dataset are different, we first retarget all motion to a chosen skeleton as in \cite{holden2016deep}. The skeleton contains 24 joints, we first extract the X and Z global coordinate of the root, and rotate the human pose to the Y-axis direction as in \cite{holden2015learning}, the global position and angle of human pose can be recovered from the X-Z velocity and the rotation velocity around the Y axis. Finally the original human pose vector contains 76 degrees of freedom, 72 for 3D joint positions, 4 for the global translation and rotation velocity. 
	
	\begin{figure}[h]
		\centering
		\includegraphics[width=0.8\linewidth]{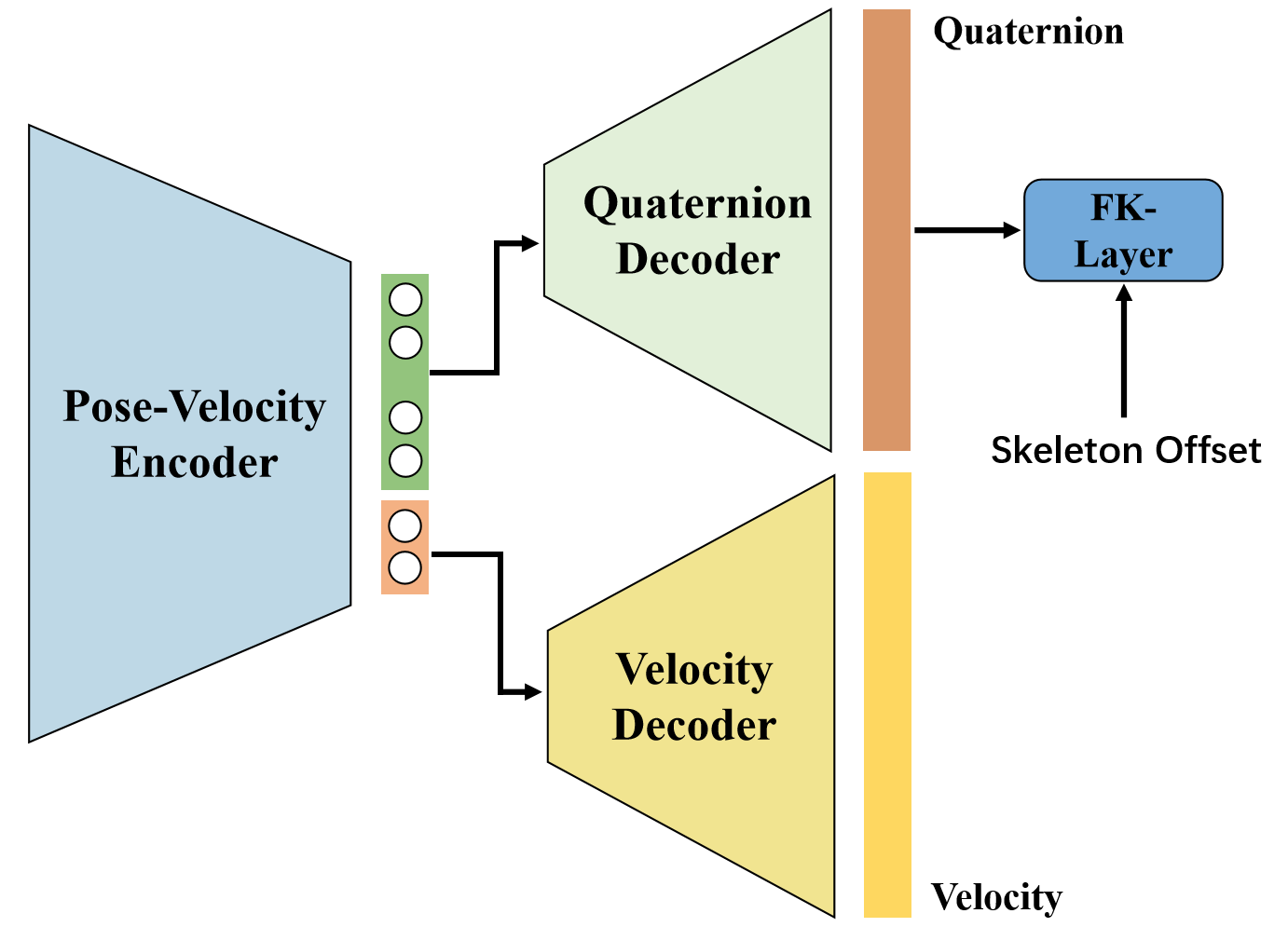}
		\caption{The pose-velocity auto-encoder network. The input of the encoder is the 76 dimensional pose-velocity vector. The encoder outputs two code, one for pose, the other for velocity. The pose code and velocity code is fed into a quaternion decoder and velocity decoder separately. The 3D joint positions are recovered from quaternions by the Forward Kinematics (FK) layer. }
		\label{fig:pose-autoenc}
	\end{figure}

	\section{Methodology}
	Formally, we start by describing a motion as a homogeneous time series: $\{X_0,\dots, X_T\}$, where $X_t$ is the motion frame at time $t$ and contains the joint positions and global velocities. Starting with a joint distribution $P(X_{<t}, X_t, X_{t+1:t+H})$, we model the influence of the past frames $X_{<t}$ and future frames $X_{t+1:t+H}$ on the current frame $X_t$ by transition probabilistic distributions $P(X_t | X_{t+1:t+H})$ and $P(X_t | X_{<t})$, where $H$ is the duration of a short-horizon future. The key reason of such a modelling choice is based two observations: the current frame is a result of the past motion and therefore conditioned on it, captured by $P(X_t | X_{<t})$. Meanwhile, the current frame is also a choice made for certain planned future, e.g. needing to stop swing the legs aiming for a transition from walk to standing, captured by $P(X_t | X_{t+1:t+H})$. In addition, since the past motion will also limit the possibilities of the future motion, there is also a impact of the past on the future, $P(X_{t+1:t+H} | X_{<t})$. Overall, the joint probability:
	\begin{equation}
	P(X_{<t}, X_t, X_{t+1:t+H}) \propto P(X_t | X_{t+1:t+H})P(X_{t+1:t+H},X_t | X_{<t})
	\end{equation}
	Not that the two probabilities on the right side play different roles. $P(X_{t+1:t+H},X_t | X_{<t})$ is the probability of unrolling from the past to the future. Given a known past, this is a joint probability of both the current and the future, containing all the possible transitions. On top of it, $P(X_t | X_{t+1:t+H})$ dictates that if the future is also known, then the current can be inferred. This \textit{explicit} modelling of the transition probabilistic distributions between the past, current and future helps capturing the transition stochasticity, which facilitates diversified motion generation as shown in the experiments.
	
	Learning the transitional probabilities in the data space, however, is difficult due to the curse of dimensionality. We therefore project motions to a latent space, which involves embedding the frames as well as the dynamics. We then learn the transition distributions in the latent space. During inference, we then recover motions from sampled states in the latent space to the original data space. DFN is naturally divided into three components: Spatial (frame) embedding, dynamics embedding and dynamics modelling.

	\subsection{Spatial Embedding}
	We use an auto-encoder for frame embedding, $z_t = PoseEnc(X_t)$ and $\hat{X}_t = PoseDec(z_t)$, shown in Figure \ref{fig:pose-autoenc}. $PoseEnc$ is multi-layer perceptron network to project the data into the latent space. Then we separate the latent feature into two components to represent the pose code and the global velocity code. $PoseDec$ contains two components, the quaternion decoder and the velocity decoder. The quaternion decoder takes the pose latent feature as input and outputs joint angles (represented by quaternions), and the velocity decoder takes the latent velocity feature as input and outputs the velocity. The quaternion decoder essentially is a differential Inverse Kinematics module. As stated in \cite{quaterNet}, using joint rotations instead of joint positions maintains the bone lengths. After the reconstruction, we use a Forward Kinematics layer to compute the 3D joint positions. To train the auto-encoder, we use a Mean Squared Error loss function:
	\begin{equation}
	\label{eq:SL}
	L_{sl} = \frac{1}{T}\sum_{t=0}^T||X_{t} - \hat{X}_{t}||^{2}_2
	\end{equation}
	where $T$ is the number of frames in a motion.
	
	\subsection{Dynamics Embedding}
	\begin{figure}[h]
		\centering
		\includegraphics[width=0.8\linewidth]{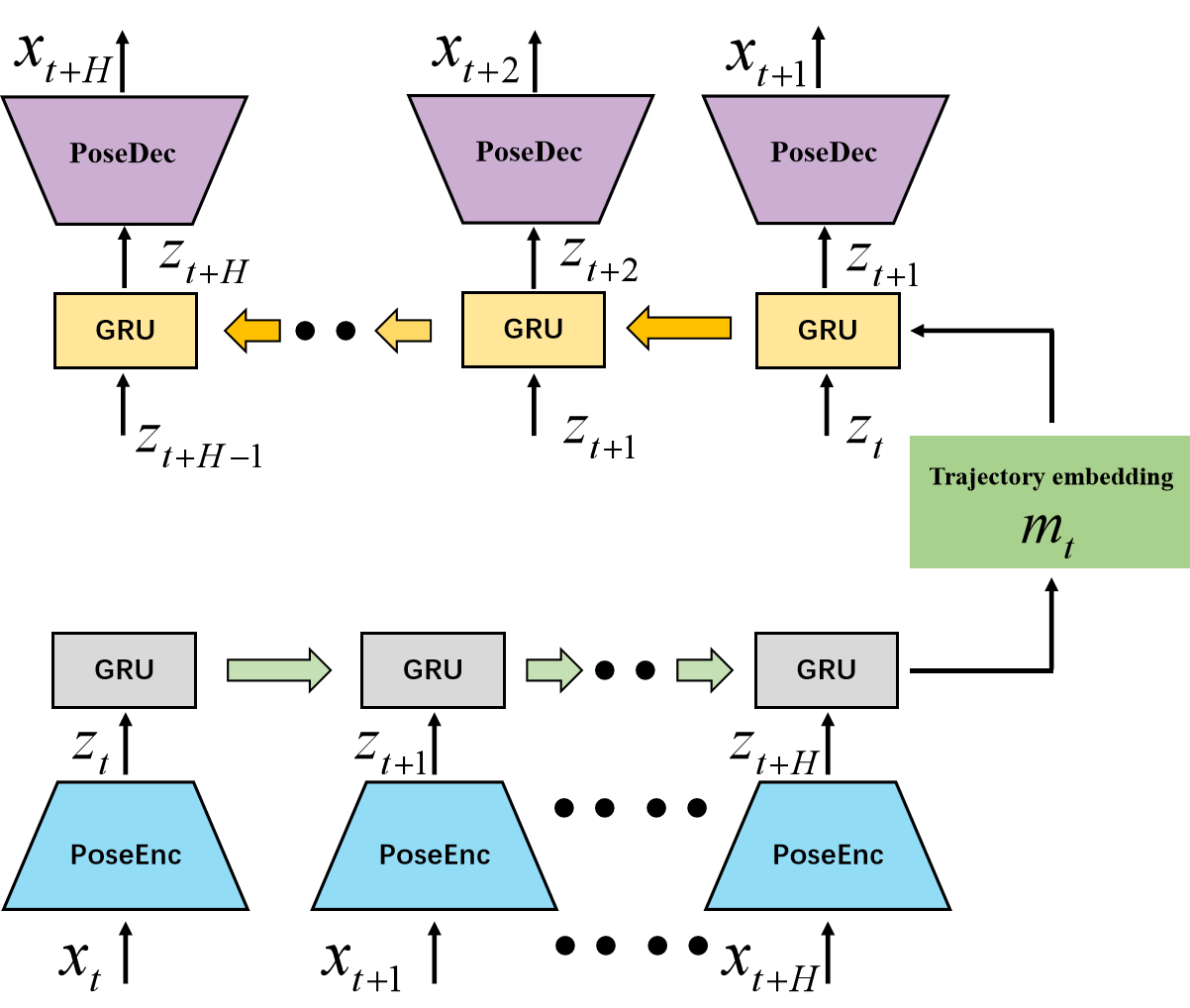}
		\caption{A seq-to-seq network for trajectory embedding.}
		
		\label{fig:traj-autoenc}
	\end{figure}

	After learning the pose latent space, we project the dynamics as trajectories in this space using a Recurrent Neural Network shown in Figure \ref{fig:traj-autoenc}. We employ a sequence-to-sequence architecture as it forces the model to learn long-term dynamics. The RNN consists of Gated Recurrent Unit (GRU)~\cite{GRU}, encodes a sequence of encoded frames $\{z_t,\dots,z_{t+H}\}$ into a latent representation $m_t$, and then unroll to reconstruct the same sequence $\{z'_{t+1},\dots,z'_{t+H}\}$ from $m_t$ given $z_{t}$. To $z_{t} = PoseEnc(X_{t})$, $m_t$ is a future summary over multiple frames. We use the following loss function:
	\begin{align}
		&L_{tl} = L_{rec} + L_{smooth} \\
		where \ \ \ \ &L_{rec} = \frac{1}{T}\sum_{t=0}^T||z_t - z'_t||_2^2 \nonumber \\ 
		&L_{smooth} = \frac{1}{T}\sum_{t=0}^T||V_t - \hat{V}_t||_2^2 \nonumber
	\end{align}
	where $T$ is the frame number of a motion, $V_t$ and $\hat{V}_t$ are the original and reconstructed joint velocities. To facilitate training, we use Eq. \ref{eq:SL} to pre-train the posture auto-encoder an fix its weights when training the RNN module.
	
	\subsection{Dynamics modelling}
	
\begin{figure*}[t]
		\centering
		\includegraphics[width=\linewidth]{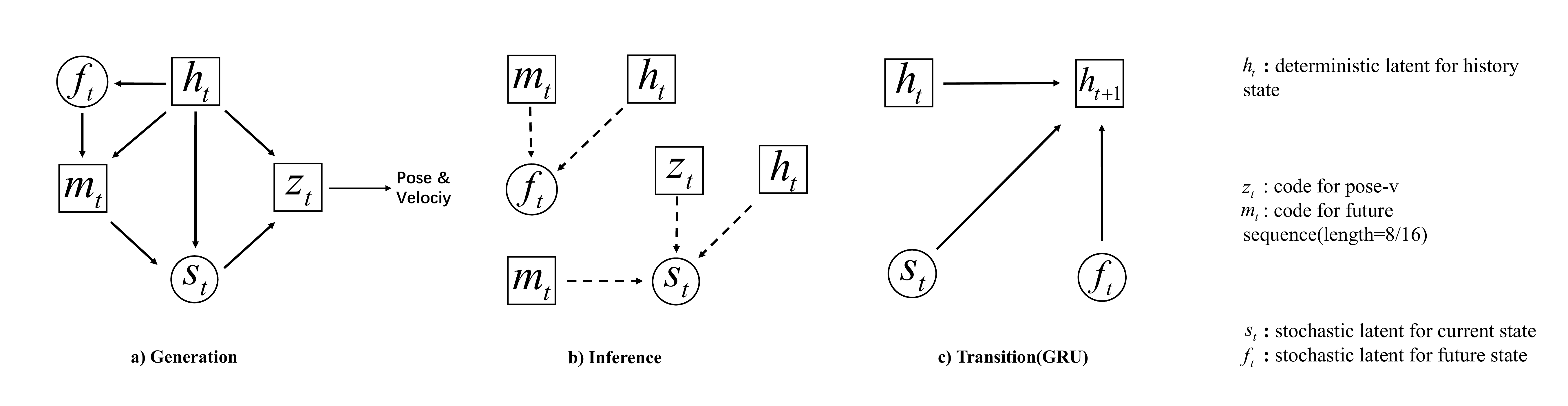}
		\caption{The stochastic latent RNN. a) Generation Model. The current pose embedding feature $z_t$ depends on the current and past latent state $s_t$ and $h_t$. $s_t$ depends on the past state $h_t$ and the future summary $m_t$ which depends on the future latent state $f_t$ and past state $h_t$ b) Inference on $s_t$ and $f_t$. c) Transition of $h_t$}
		\label{fig:latent-rnn}
	\end{figure*}

	\textbf{Generative Model.} After embedding the poses and dynamics into a latent space, we now explain the dynamics modelling, which is the key technical contribution of this paper.  We propose a new dynamics model that captures the joint distribution $P(X_{<t}, X_t, X_{t+1:t+H})$. Rather than directly learning the distribution in the data space, we aim to learn the latent joint distribution $P(z_{<t}, z_t, z_{t+1:t+H})$, where we abstract the past, current and future features separately. First, given the Markov property, we assume that all past information $z_{<t}$ is encoded into $h_t$ which is a deterministic (known) past state. Next, we assume that the future information $z_{t+1:t+H}$ can be summarized into a future state $f_t$, and $f_t$ is drawn from a distribution over all possible future states conditioned on $h_t$. Last, we also assume that there is a current state $s_t$ which captures the current information $z_t$ and $s_t$ is drawn from a distribution of all possible current states. Then we can therefore assume:
	\begin{align}
		P(z_{<t}, z_t, z_{t+1:t+H}) &= P(z_{t+1:t+H},z_t | z_{<t}) P(z_{<t}) \nonumber \\
		&\propto P(z_t | z_{t+1:t+H})P(z_t | z_{<t})P(z_{t+1:t+H} | z_{<t}) \nonumber \\
		&= P(s_t | f_t)P(s_t | h_t)P(f_t | h_t) \nonumber \\
		&= P(s_t | f_t, h_t)P(f_t | h_t)
	\end{align}
	where we directly use $s_t$, $h_t$ and $f_t$ to replace the corresponding $z$ variables by assuming mappings between them which will be explained later. Different from existing methods \cite{goyal2017z,serdyuk2018twin}, our direct conditioning of the current state on the future and past state $P(s_t | f_t, h_t)$, and the future state on the past state $P(f_t | h_t)$ allows us great flexibility in modelling the stochasticity in transitions. The generation model is shown in Fig.\ref{fig:latent-rnn} a.
	
	\textbf{Future Feature Prior.}  Given $h_t$, we first predict the future state, via $P(f_t | h_t)$. Here, we assume a diagonal multivariate Gaussian prior over $f_t$ \cite{chen2016dynamic,habibie2017recurrent,karl2017deep} 
	%[any reference to say we can assume that this is a Gaussian??????]:
	\begin{align}
		p(f_{t} | h_{t}) = Gaussian( f_{t}; \mu^{f}_{t}, \sigma^{f}_{t} ), \nonumber  \\
		\text{where} \ \ [\mu^{f}_{t}, \sigma^{f}_{t}] = g^p_f(h_{t})
	\end{align}
	where $\mu^{f}_{t}$ and $\sigma^{f}_{t}$ are the mean and covariance. $g^p_f$ is a three-layer MLP with hidden dimension 256 and LeakyReLU activation. It contains all the possible future state given the past. It can represent a goal or a driving signal for the generative process. Also it forces the model to learn rich motion transitions and long term correlations, overcoming the freezing problem of traditional RNN \cite{zhou2018autoconditioned}.
	
	\textbf{Current Feature Prior.} Next, we explain $P(s_t | f_t, h_t)$. Although $h_t$ is a known (deterministic) past, $f_t$ is random. We therefore first sample a specific future state $f_t$, then decode it to an unrolled future summary $m_t$, and finally condition the current state $s_t$ on $h_t$ and $m_t$. We therefore have:
	\begin{align}
		P(s_t | f_t, h_t) \propto P(s_t | m_t, h_t)P(m_t | f_t, h_t) \\
		P(s_t | m_t, h_t) = Gaussian(s_t; \mu^s_t,\sigma^s_t) \\ 
		[\mu^s_t, \sigma^s_t] = g^p_s(h_t, m^p_t) \nonumber
	\end{align}
	where $P(m_t | f_t, h_t)$ is parameterized by a  four layer MLP with hidden dimension 128 with LeackyReLU activation. $\mu^{s}_{t}$ and $\sigma^{s}_{t}$ are the mean and covariance. $g^p_s$ is a two layer MLP with hidden dimension 128. After being able to sample the current state $s_t$, we can compute the current feature $z_t$ via $z_{t} = MLP(s_{t}, h_{t})$, where the MLP has three hidden layer with 128 dimensions and LeakyReLU activation.  
	
	Finally, given the current and future state, the past state is updated as follows (Fig.\ref{fig:latent-rnn} c):
	\begin{equation}
	h_{t+1} = \mathbf{GRU}(h_{t}, s_{t}, f_{t})
	\end{equation}
	where the GRU has two stacked layer and hidden state of dimension 128. Now the generation model in Fig.\ref{fig:latent-rnn} a is complete.
	
	Different from existing methods \cite{goyal2017z,chung2015recurrent} where the prior of current state is a function of past state $h_{t}$ only, and where the future state is shared with the current state, we let the model learn two different distributions for current and future states. The prior of current state is also a function of the future state, which will force the model to make use of the future information.

	\subsection{Inference} 
	In the generation model in Fig.\ref{fig:latent-rnn} a, the key variables to be inferred are $s_t$ and $f_t$, shown in Fig.\ref{fig:latent-rnn} b.
	The posterior of the future state $f_t$ is dependent on past state $h_t$ and its unrolled future summary $m_t$. The posterior of the current state $s_t$ is dependant on the feature $z_t$, the past state $h_t$ and the future summary $m_t$. We first factorize the dynamics as follow:
	\begin{equation}
	q(s_{\leq T}|z_{\leq T}) \approx \prod _{t=1}^{T}q(s_{t}|z_{\leq t-1}, z_t,  z_{\leq t+H}) = \prod _{t=1}^{T}q(s_{t}|h_t, z_{t}, m_{t})
	\end{equation}
	where $T$ is the total length of motion. Here we approximate the $q(s_{t} | z_{\leq T} )$ with $q(s_{t} | z_{\leq t+H})$, as the correlations between $s_t$ and the far future is likely to be small, so we only consider up to $t+H$. Then for each time step, we use a MLP to parameterize the posterior:
	\begin{equation}
	q(s_{t}|h_{t}, z_{t}, m_{t}) = Gaussian(\mu_{t}^{s}, \sigma_{t}^{s}),[\mu_{t}^{s}, \sigma_{t}^{s}]=MLP(h_{t}, z_{t}, m_{t})
	\end{equation}
	where the MLP has two hidden layers with 32 dimensions and LeakyReLU activation. For the future state we approximate its posterior as follow:
	\begin{equation}
	q(f_{t}|h_{t}, m_{t}) = Gaussian(\mu_{t}^{f}, \sigma_{t}^{f}),[\mu_{t}^{f}, \sigma_{t}^{f}]=MLP(h_{t}, m_{t})
	\end{equation}
	where the MLP has two hidden layers with 512 dimensions and LeakyReLU activation.

	\subsection{Temporal difference loss}
	Besides inferring $s_t$ and $f_t$, we also constrain the dynamics of $s$. We assume a relation between two states $s_{t_1}$ and $s_{t_2}$ at $t_1$ and $t_2$ where $t_1 < t_2$, similar to \cite{gregor2018temporal}:
	\begin{align}
		p(s_{t_1}, s_{t_2}, t_1, t_2) \propto p(s_{t_2} | s_{t_1}, h_{t_1}, h_{t_2})
	\end{align}  
	where we parameterize the posterior:
	\begin{align}
		q(s_{t_1} | s_{t_2}, h_{t_1}, h_{t_2}) =  Gaussian(s_{t_1}; \mu_{s_{t_{1}}}, \sigma_{s_{t_{1}}}) \\
		\text{where}\ \ \mu_{s_{t_{1}}}, \sigma_{s_{t_{1}}} = MLP(s_{t_{2}}, h_{t_{1}}, h_{t_{2}}) \nonumber
	\end{align}
	where $\mu_{s_{t_{1}}}$ and $\sigma_{s_{t_{1}}}$ are the mean and covariance. Here the MLP has two hidden layers with 32 dimensions and LeakyReLU activation. This way we can sample $s_{t_{1}}^{posterior}$ during inference. Meanwhile, we hope to reconstruct $s_{t_2}$ with the time difference $\delta t = |t_{1} - t_{2}| $
	\begin{equation}
	s_{t_{2}}^{rec} = MLP(s_{t_{1}}^{posterior},\delta t)
	\end{equation}
	where the MLP for skip prediction has three hidden layers each with 32 dimensions and LeakyReLU activation.

	\subsection{Learning} 
	Finally, we compose all terms for the loss funtion:
	\begin{align}
		max_{\Phi} \ \ L_{d} + L_{T}
	\end{align}
	where $\Phi$ is the set of all learnable parameters in our networks and
	\begin{align}
		L_{d} = \sum_{t=1}^T[-KL ( q( s_t | h_t, z_t, m_t) || p( s_{t} | h_t, m_{t}) ) \nonumber \\
		- KL( q( f_{t} | h_t, m_t ) || p( f_t | h_t  )  ) \nonumber \\
		+ log(p(z_{t} | s_t, h_t)) + log(p(m_t|f_t, h_t))] \nonumber \\
		L_{T} = \sum_{t_{1} ,t_{2}}[-KL( q(s_{t_{1}} | s_{t_{2}}, h_{t_{1}}, h_{t_{2}} ) ||p(s_{t_{1}}) ) + p(s_{t_{2} }|s_{t_{1} }, t_{1}, t_{2})] \nonumber
	\end{align}
	$KL$ is the Kullback–Leibler divergence.
	
	After training the pose auto-encoder and the sequence auto-encoder (Section 5.1-5.2), we freeze their parameters and train the dynamics model (Section 5.3).

	\section{Experiment and Results}
	For all our experiments, we use CMU MoCap database\footnote{\href{http://mocap.cs.cmu.edu/}{http://mocap.cs.cmu.edu/}}. CMU dataset is a high-quality dataset acquired using optical motion capture systems, containing 2605 trials in 6 categories and 23 subcategories. Its high-quality serves our purposes well as it provides good data `seeds' for motion generation. Also, the tremendous effort that went into capturing the data shows the need for tools such as DFN for data augmentation. To carefully evaluate DFN, we select different motion classes with different features and dynamics, shown in Tab.\ref{tab:motionType}, to show that DFN can generate new high-quality motions with arbitrary lengths using different motion prefixes. Next, we evaluate DFN on data with single and mixed motion classes to see its ability to learn the different transition stochasticity on data with a single type of dynamics and mixed types of dynamics. Last, we push the limit of DFN by reducing the training data, to show that DFN can make use of a small amount of data to generate high-quality and diversified data, which is crucial for data augmentation. More example can be found in the supplementary video.

	\subsection{Open-loop Motion Generation}
	We first show open-loop motion generation, where we do not moderate accumulative errors.  We use a 8 to 20-frame motion prefix to start motion generation to get 900 frames (dfn\_run2box\_2char and dfn\_boxing\_3char in the video). The motion stability indicates that DFN does not suffer from the problem of cumulative error that is common in time-series generation \cite{wang2019spatio}. Given the same prefix, the diversity is shown in their transitions between different postures (short-term) and different actions (long-term). 
	
	\subsection{Dynamics Multi-modality}
	We investigate how well DFN can capture different transition stochasticity in different motions, using several types of motions with different properties (shown in Tab.\ref{tab:motionType}). We first train DFN on them separately then jointly. The results can be found in dfn\_walk\_top, dfn\_walk\_close, walking1-walking2, running1-running3, dancing1 and boxing1 in the video. We observe that DFN can learn the transition stochasticity well when trained on single type of motions. The diversity can be found in short-term and long-term transitions, which are two-levels of multi-modality captured well by DFN. In walking (dfn\_walk\_top and dfn\_walk\_close in video), the short-term stochasticity is shown in within-cycle motion randomness, which enriches the walking style. The long-term stochasticity is shown when a turning is generated. The action-level transition has also been captured and generated. Similar observations are also found in other motions. 
	
	When trained on mixed data (combining all motion in Tab.\ref{tab:motionType}), DFN learns higher-level action transitions between different motion classes. We can see examples (action\_transition in video and frame-level image in supplementary file) that transit from dancing to running, from boxing to dancing or from slow walk to running, showing the modelling capacity at two levels. Within a single action, diversified styles are learned well. Between different actions, transitions are learned well too. This demonstrates the benefits of modelling randomness explicitly between the past, current and future state, which would otherwise make it hard to capture the multi-modality and lead it to average over all types of dynamics, resulting in meaningless mean poses and motions.
	
	\begin{table}[tb]
		\centering
		\begin{tabular}{c|c|c|c|c}
			Motion&Cyclic&Main Body Part&Rhythmic&Dynamics  \\
			\hline
			Walking&Yes&Lower&No&Low \\
			\hline
			Boxing&No&Upper&No&High \\
			\hline
			Dancing&No&Full&Yes&High \\
			\hline
			Running&Yes&Lower&No&High
		\end{tabular}
		\caption{Motion types and their features.}
		\label{tab:motionType}
	\end{table}
	% \vspace{-1.5em}

	\subsection{Diversified Generation} 
\begin{figure}[tb]
		\centering
		\includegraphics[width=0.8\linewidth]{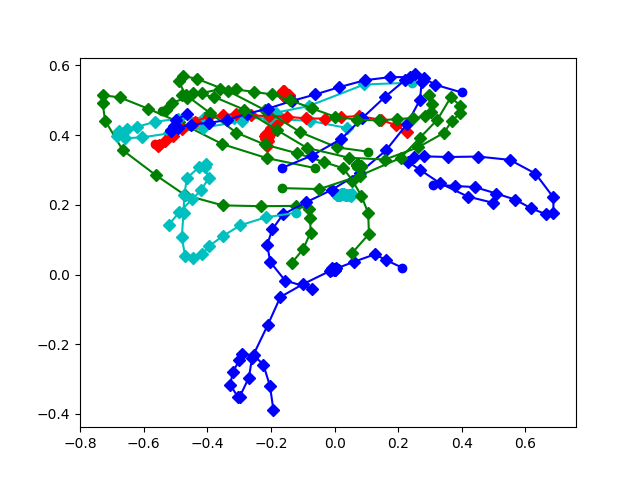}
		\caption{Randomly selected training motions in the latent space. Color indicate different motion class. Smooth trajectories are universally obtained by embedding.}
		\label{fig:project_train_motion}
	\end{figure}

\begin{figure}[tb]
		\centering
		\includegraphics[width=0.8\linewidth]{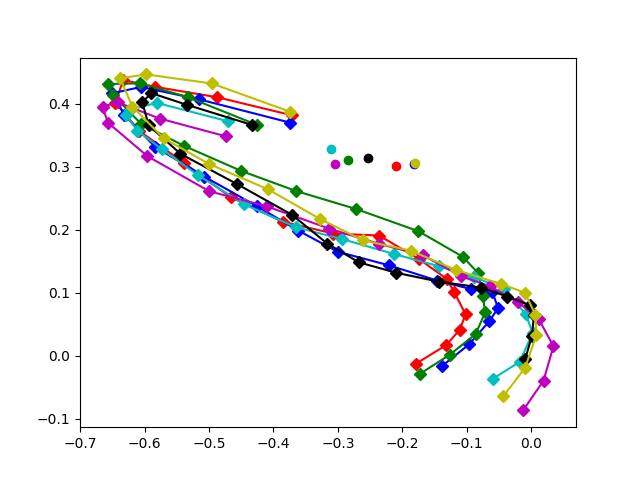}
		\caption{Pose embedding trajectory of random generated sequences given same initialization with 20 frame. The circle mark the first frame. We see that as time goes, these trajectory depart from each other.}
		\label{fig:generate_trajectory_same_init}
	\end{figure}

	Although visually it is clear that DFN can generate diversified motions, we also numerically show the diversity especially when the duration of generation becomes long. First, we randomly select training data of different classes, and show their latent feature trajectory in Fig. \ref{fig:project_train_motion}, with embedding dimension of 16. We then use a PCA model to project the embedding features to 2D. The trajectories are continues and smooth without extra constraints on the auto-encoder. It shows that the motion dynamics are captured well in the latent space, which is critical in motion generation. Next, we show a group of generated motions in Fig. \ref{fig:generate_trajectory_same_init}. Even with the same motion prefix, motions start to diversify from the beginning, which is a distinct property lacking in deterministic generators in most action-prediction models such as \cite{martinez2017human}, and our sequence is in 3d which is more difficult than 2d \cite{Diverse-Stochastic-Human-Action}.
	
\begin{figure}[tb]
		\centering
		\includegraphics[width=\linewidth]{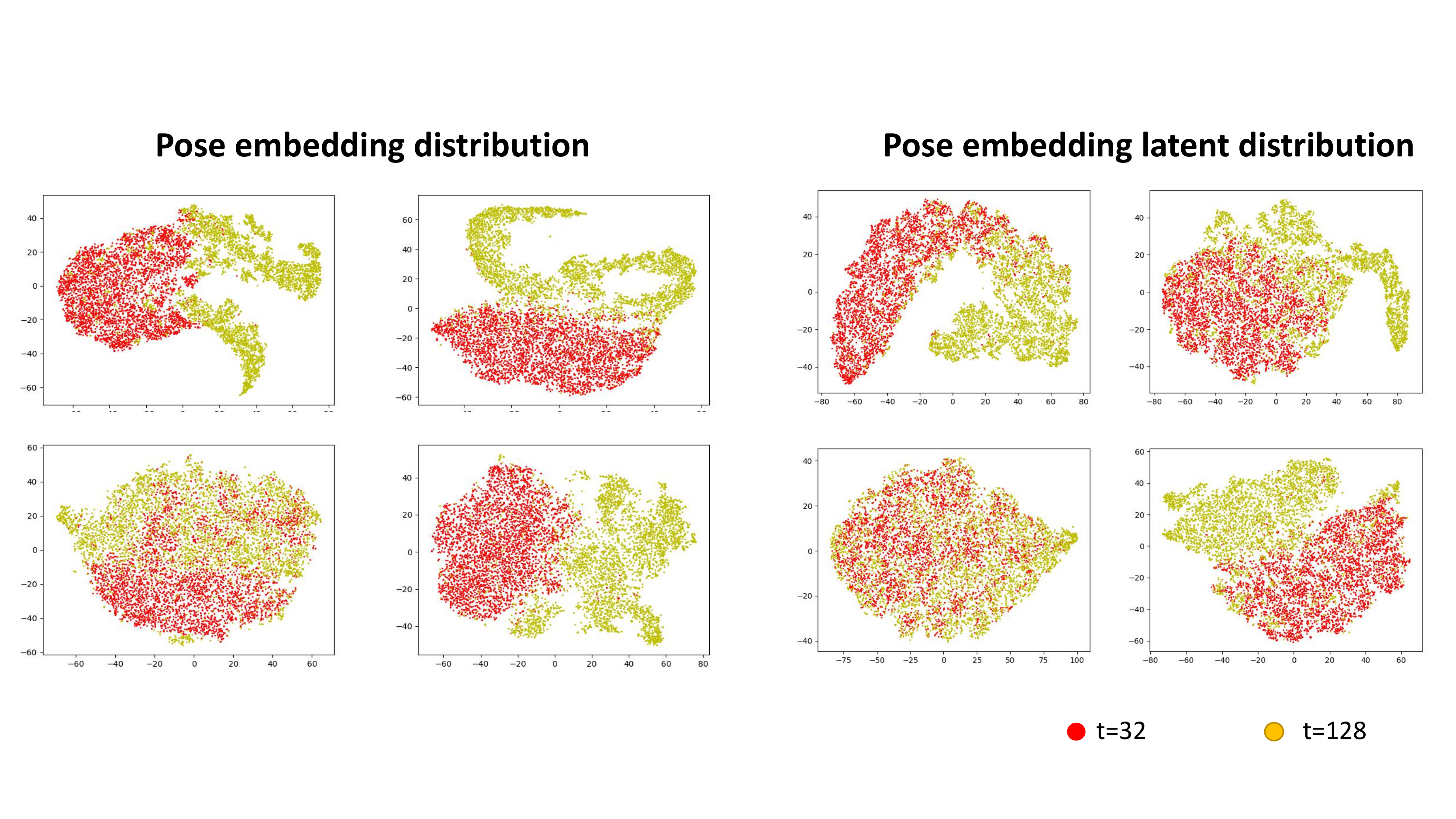}
		\caption{Four groups of motions generated from four different motion prefixes, each group with 4096 motions, and their $z$ (Left) and $s$ (Right) at $t=32$ and $t=128$. We can see that the earlier distributions are more concentrated and diverge fast as time passes.}
		\label{fig:embed_latent_dist}
	\end{figure}

	\textbf{Distributional Shift in Time.} The motion diversity increases in time. To show that there is a distributional shift of poses, we randomly pick an initial sequence from training data, then randomly generate 4096 sequences each with 128 frames. We visualize the current latent state $s$ and latent feature $z$ at $t=32$ and $t=128$ in Fig. \ref{fig:embed_latent_dist}. Note that the distributions of $s$ and $z$ capture the stochasticity at two different levels, one at the stochastic state level and the other at the latent feature level. The red dots represent them at $t=32$ and the yellow dots at $t=128$. 
	
	For both $z$ and $s$, the red ($t=32$) are concentrated more, showing that the difference between the 4096 generated motions are still somewhat similar in the early stage. However, the yellow ($t=128$) show that the generated motions start to diversify later. Not only they shift out of the original red region, indicating that they are in now in different pose regions, they also start to diverge more, shown by different modes in yellow areas, meaning they have diverged into several different pose regions.

	\begin{figure}[tb]
		\centering
		\includegraphics[width=\linewidth]{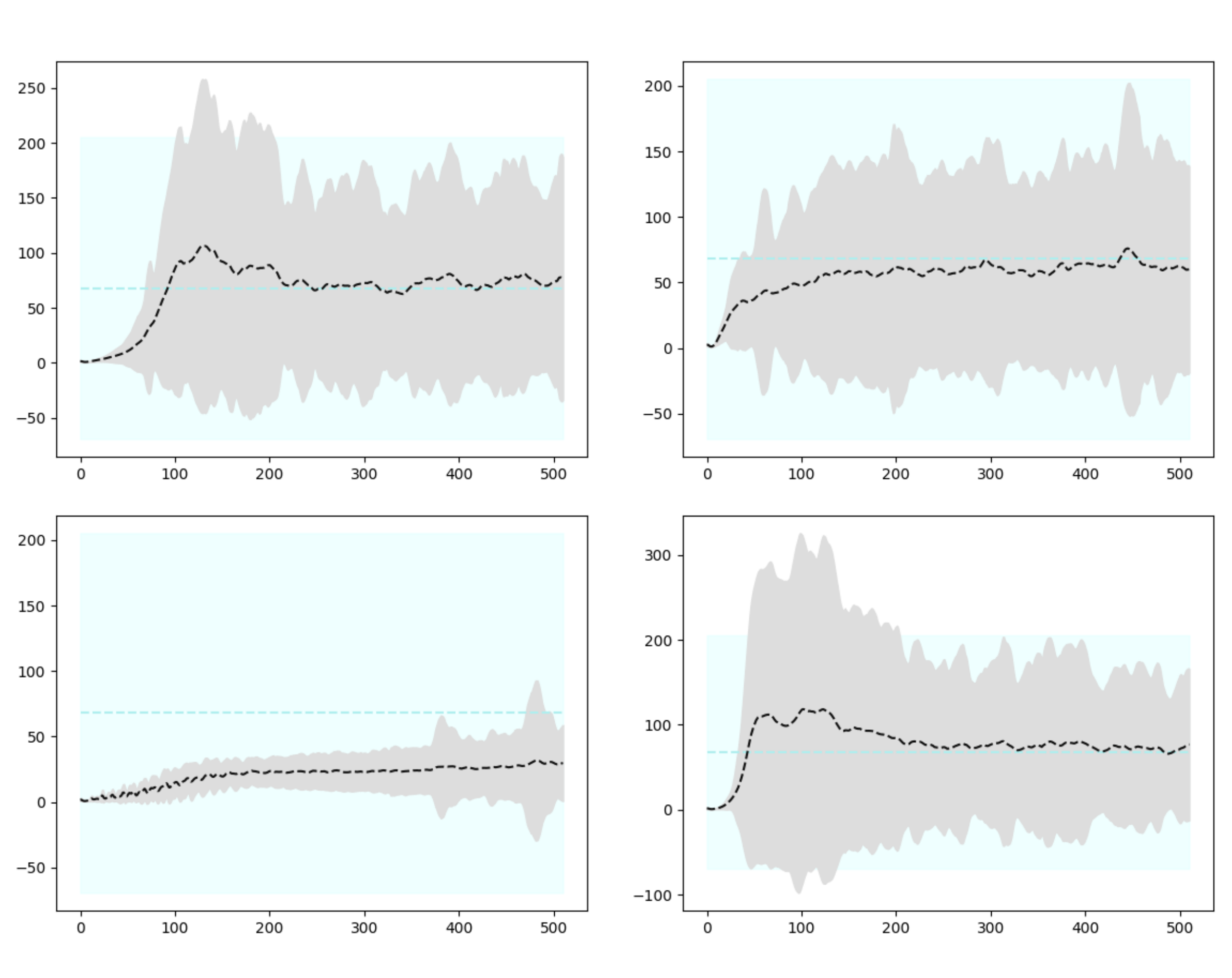}
		\caption{Four groups of motions generated from four different motion prefixes, each group with 4096 motions. The x axis represents the time dimension, and the y axis represent the mean distance to average pose at each time step. The band represents the variations.}
		\label{fig:mean_pose_dist}
	\end{figure}
	
	\textbf{Distribution Matching in Time}. Another way to test the diversity of generated motions is to see their statistical similarity to the training motions. Since the motion prefix is from one particular motion, the more similar the generated motions are to the whole training dataset, the more diverse they are, because the generated motions have leave the original motion region where the motion prefix is. 
	
	We employ the mean-distance distribution as a measure, as in \cite{Diverse-Stochastic-Human-Action}. For each time step, we calculate the mean pose of all generated motions, then calculate the Euclidean distances between the mean pose and all other poses at that time step. We then plot the mean distance and variance in Figure \ref{fig:mean_pose_dist}. The blue background indicate the mean and variance of mean-distance distribution of the training dataset. It shows that as time goes, the mean-distance distribution of generated poses gradually matches that of the training data. This further shows the generation diversity. 
	
	\subsection{Generation on Limited Training Data}
	DFN aims to solve the problem of data scarcity, so it should only require as little data as possible for generation. We therefore push DFN to its limit by reducing the training data, to see the minimal amounts of data needed. To investigate each individual type of motions, we train DFN on walking, running, and boxing data separately. We start from full training data where the longest sequence lasts for around 10 minutes, and gradually reduce the duration by sampling until the quality of the generated motions start to deteriorate. Although DFN responds to reduced training data slightly differently on different motions, we finally able to reduce the training data to a tiny amount, with the longest sequence being only 15 seconds (12 second for walking, 15 second for boxing and 7 second for running). DFN can still generate stable motions even when trained on merely a 7-second long motion. (The result can be seen in reduced\_data in video) The impact of reducing the training data is mainly on the diversity of the motion. (However we can see in supplementary video that the generated boxing motion still has a certain of diversity). Less training data contains fewer transition diversities (both short-term and long-term). The generated motions therefore are less diverse. This is understandable as DFN cannot deviate too much from the original data distribution to ensure the motion quality.
	
	\subsection{Comparison}
	To our best knowledge, the only similar paper to ours is \cite{wang2019combining} which also focuses on diversified motion generation. However, the biggest difference is that DFN explicitly models the influence of the future on the current. This enables DFN to explicitly model the transition randomness at different stages and levels. This is the key reason why DFN can be trained well on multiple types of motions, separately and jointly, which has not been shown in \cite{wang2019combining}. However, a direct numerical comparison is difficult due to the lack of widely accepted metrics for diversified motion generation. In addition, the method in \cite{wang2019combining} uses heavy post-processing while DFN does not.

	\section{Conclusion and discussions}
	In this paper, we propose a new generative model, DFN, for diversified human motion generation. DFN can generate motions with arbitrary lengths. It successfully captures the transition stochasticity in short and long term, and capable of learning the multi-modal randomness in different motions. The training data needed is small. We have conducted extensive evaluation to show DFN's robustness, versatility and diversity in motion generation.
	
	There are two main limitations in our method. There is no control signal, and sometimes it can overly smooth high-frequency motions. We will address them in the future. Our explicit modelling of the future makes it convenient to introduce desired future as control signals;while replacing some of the Gaussian components with multi-modal priors might mitigate the over-smoothing issue.

	\begin{acks}
		We thank anonymous reviewers for their valuable comments. This work is partially supported by the National Key Research \& Development Program of China (No. 2016YFB1001403), NSF China (No. 61772462, No. 61572429, No. U1736217), the 100 Talents Program of Zhejiang University, Strategic Priorities Fund Research England, and EPSRC (Ref:EP/R031193/1).
	\end{acks}

	\bibliographystyle{ACM-Reference-Format}
	\balance
	\bibliography{ref}

\end{document}